\documentclass[sigconf, nonacm]{acmart}

\def\BibTeX{{\rm B\kern-.05em{\sc i\kern-.025em b}\kern-.08emT\kern-.1667em\lower.7ex\hbox{E}\kern-.125emX}}
    
\copyrightyear{2019} 
\acmYear{2019} 
\setcopyright{acmcopyright}
\acmConference[PEARC '19]{Practice and Experience in Advanced Research Computing}{July 28-August 1, 2019}{Chicago, IL, USA}
\acmBooktitle{Practice and Experience in Advanced Research Computing (PEARC '19), July 28-August 1, 2019, Chicago, IL, USA}
\acmPrice{15.00}
\acmDOI{10.1145/3332186.3333046}
\acmISBN{978-1-4503-7227-5/19/07}

\usepackage{listings}
\usepackage{xcolor}  
\definecolor{codebg}{rgb}{0.96,0.96,0.96}
\definecolor{codegreen}{rgb}{0,0.6,0}
\definecolor{codegray}{rgb}{0.5,0.5,0.5}
\definecolor{codepurple}{rgb}{0.58,0,0.82}
\definecolor{backcolour}{rgb}{0.95,0.95,0.92}

\lstnewenvironment{onelinerSQL}{
\lstset{
    language=SQL,
    backgroundcolor=\color{codebg},
    extendedchars=true,
    basicstyle=\footnotesize\ttfamily,
    showstringspaces=false,
    showspaces=false,
    tabsize=2,
    breaklines=true,
    showtabs=false
}}{}

\lstdefinestyle{snippetstyle}{
    frame=tb,
    backgroundcolor=\color{codebg},   
    commentstyle=\color{codegreen},
    keywordstyle=\color{magenta},
    numberstyle=\tiny\color{codegray},
    stringstyle=\color{codepurple},
    basicstyle=\footnotesize\ttfamily,
    breakatwhitespace=false,         
    breaklines=true,                 
    captionpos=b,                    
    keepspaces=true,                 
    numbers=left,                    
    numbersep=5pt,                  
    showspaces=false,                
    showstringspaces=false,
    showtabs=false,                  
    tabsize=2,
    basewidth = {.50em},
    aboveskip={5mm},
    belowskip={5mm},
}

\begin{document}

\title[Shuffler: A Large Scale Data Management Tool for ML in Computer Vision]{Shuffler: A Large Scale Data Management Tool for Machine Learning in Computer Vision}

%
\author{Evgeny Toropov}
\affiliation{%
  \institution{Carnegie Mellon University}
  \streetaddress{5000 Forbes Ave}
  \city{Pittsburgh}
  \state{Pennsylvania}
  \postcode{15289}
}
\email{etoropov@andrew.cmu.edu}

\author{Paola A. Buitrago}
\affiliation{%
  \institution{Pittsburgh Supercomputing Center}
  \institution{Carnegie Mellon University}
  \city{Pittsburgh}
  \state{Pennsylvania}
  \country{USA}
}
\email{paola@psc.edu}

\author{Jos\'e M. F. Moura}
\affiliation{%
  \institution{Carnegie Mellon University}
  \city{Pittsburgh}
  \state{Pennsylvania}
  \country{USA}
}
\email{moura@ece.cmu.edu}

%
\renewcommand{\shortauthors}{Toropov, Buitrago and Moura}

%
\begin{abstract}
Datasets in the computer vision academic research community are primarily static. Once a dataset is accepted as a benchmark for a computer vision task, researchers working on this task will not alter it in order to make their results reproducible. At the same time, when exploring new tasks and new applications, datasets tend to be an ever changing entity. A practitioner may combine existing public datasets, filter images or objects in them, change annotations or add new ones to fit a task at hand, visualize sample images, or perhaps output statistics in the form of text or plots. In fact, datasets change as practitioners experiment with data as much as with algorithms, trying to make the most out of machine learning models. Given that ML and deep learning call for large volumes of data to produce satisfactory results, it is no surprise that the resulting data and software management associated to dealing with live datasets can be quite complex. As far as we know, there is no flexible, publicly available instrument to facilitate manipulating image data and their annotations throughout a ML pipeline. In this work, we present Shuffler, an open source tool that makes it easy to manage large computer vision datasets. It stores annotations in a relational, human-readable database. Shuffler defines over 40 data handling operations with annotations that are commonly useful in supervised learning applied to computer vision and supports some of the most well-known computer vision datasets. Finally, it is easily extensible, making the addition of new operations and datasets a task that is fast and easy to accomplish. 
\end{abstract}

%

\begin{teaserfigure}
  \center
  \includegraphics[width=0.85\textwidth]{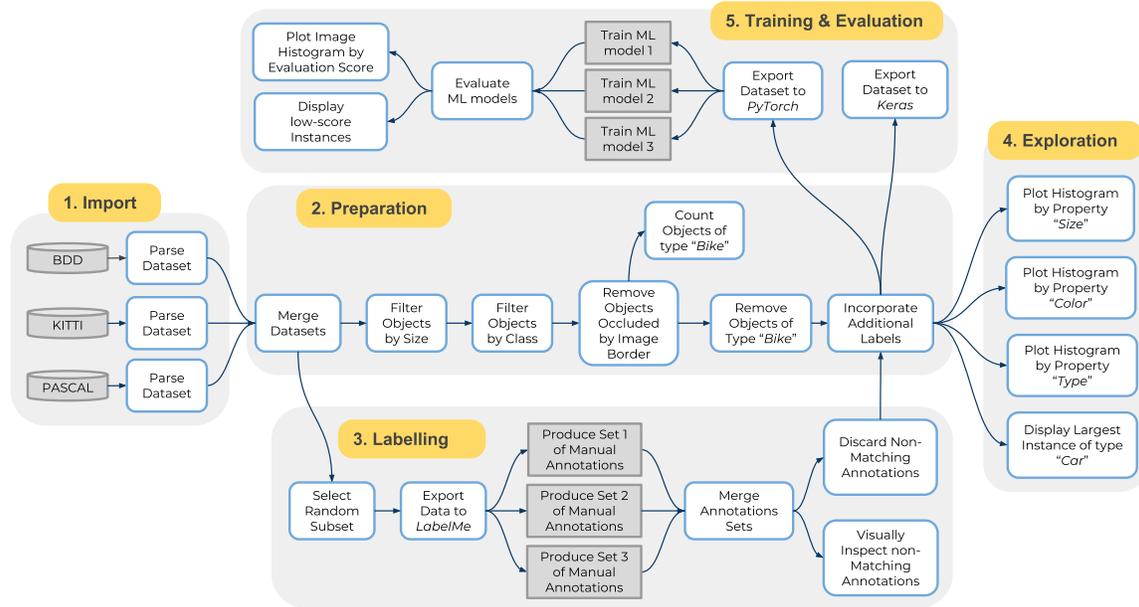}
  \caption{A simplified typical ML data workflow for object detection in computer vision. The white rectangles represent operations and the arrows show the direction in which data flows. Each operation produces new data and/or modifies the input data. Data management and software maintenance complexity is significant, and a specialized open-source data management system that simplifies operations (white rectangles) is necessary for practitioners.}
  \label{fig:teaser}
\end{teaserfigure}


%
\maketitle

\section{Introduction}

In the computer vision academic community, day-to-day work emphasizes primarily algorithms rather than data. From this point of view, annotated image datasets are ideally built once and remain fixed. This approach allows the community to use datasets as benchmarks. Researchers choose to store these datasets in formats that are most common and fast to load for machine learning (ML) packages. 
In contrast, for a data scientist in industry, the task is not necessarily to improve an algorithm, but rather to try different algorithms and tasks on various partitions and modifications of the same dataset. In this case, a dataset is not considered static, but rather constantly altered to fit the task at hand. In turn, multiple versions of the same dataset need to co-exist in a centralized or a distributed storage system. Ideally, a practitioner would want 1) a simple way to manipulate image data and its annotations, and 2) a file format that allows to store multiple copies of the annotation set in an organized and efficient way and to inspect them manually.

Data manipulation tools are sometimes packaged with a dataset, but they typically allow to perform only a limited number of operations only on that particular dataset and often for a single programming language. An example is the PASCAL VOC dataset~\cite{PASCAL_retrospective} that had $10$ releases in different years and with each release coming with a Matlab toolbox for that specific year. In alternative, researchers often write in-house small one-time scripts to quickly alter a dataset, for example, to change the size of object bounding boxes. This set of scripts usually contains duplicate code, tends to be error prone, and with time becomes increasingly difficult to maintain.

\setlength{\tabcolsep}{4pt}
\begin{table}
\caption{Raw annotation formats of popular object detection datasets are not designed for cross-data manipulation. Time benchmarks were written in Python and run on an 2.9 GHz Intel Core i5 computer with an SSD hard-drive.}
\label{table:dataset_formats}
\begin{tabular}{ccc}
\toprule
Dataset & Annotation file format & Load + parse time \\
\midrule
PASCAL2012~\cite{PASCAL_retrospective} & \texttt{xml} file per image & 1 sec \\
KITTI~\cite{2012_KITTI} & \texttt{txt} file per image & 6 sec \\
Cityscapes~\cite{Cordts16_Cityscapes} & \texttt{json} file per image & 20 sec \\
BDD~\cite{2016_bdd} & a single \texttt{json} file & 4 sec\\
\bottomrule
\end{tabular}
\end{table}
\setlength{\tabcolsep}{1.4pt}

Datasets typically come in a custom format, which usually includes images and annotation files in one of the following formats: \texttt{xml}, \texttt{txt}, or \texttt{json}. Table~\ref{table:dataset_formats} presents an overview of several popular object detection datasets in the area of autonomous driving and the formats of the associated annotation files. On the one hand, these formats are human-readable, but on the other hand, quite slow to load. Additionally, changing annotations and saving them as a copy means duplicating the whole directory with the annotation files, which is inconvenient and slow. Many development kits cache annotations by serializing them with formats such as \texttt{pickle}\footnote{\url{https://docs.python.org/3/library/pickle.html}} or \texttt{protobuf}\footnote{\url{https://developers.google.com/protocol-buffers/?hl=en}}. Such formats are easy to load by a machine learning framework and convenient to store, but they are not interpretable by humans and can not be inspected or modified outside of a specialized programming environment.

\begin{figure}
  \centering
  \includegraphics[width=0.4\textwidth]{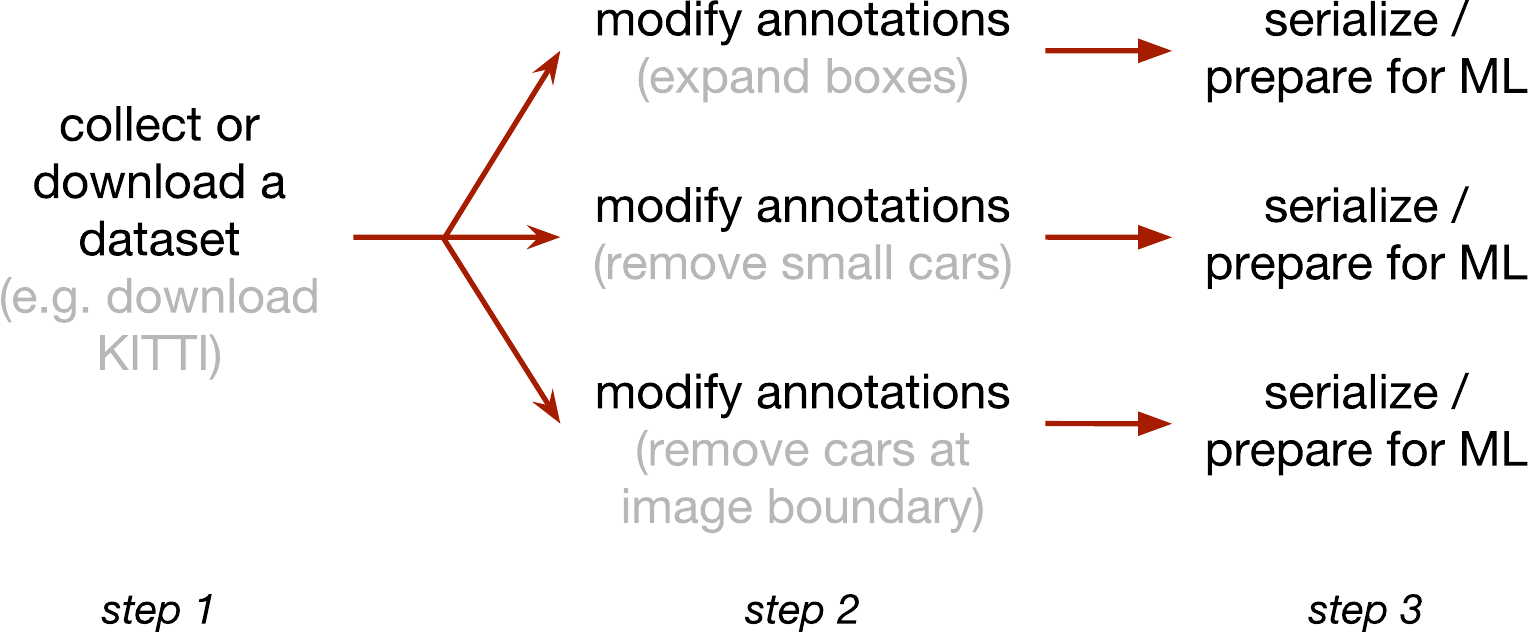}
  \caption{Typical data preparation pipeline lacks convenient tools on step $2$.}
  \label{fig:data-preparation-pipeline}
\end{figure}

To sum up, we consider a typical data preparation workflow of a computer vision practitioner to consist of three steps (Figure~\ref{fig:data-preparation-pipeline}): 1) download or collect a dataset, 2) modify annotations, and 3) serialize the dataset. We further consider a common situation when multiple modifications of annotations are used. Modifications could be a chain of trivial tasks, for example, removing objects at image boundaries and then increasing the size of bounding boxes. We note 1) the lack of software for manipulating image data and annotations, and 2) a convenient format to store annotations.

In this work, we close this gap by proposing a software toolbox, Shuffler, designed specifically for manipulating annotations. It employs widely known relational databases and the associated SQL query language for storing and manipulating annotations. The proposed toolbox is heavily based on SQL and allows to chain multiple operations in a single command. Annotations are stored in an relational database (Sqlite, MySql, ...) with schema designed to cover the bulk of the common tasks in computer vision. The proposed solution satisfies the following properties:

\begin{itemize}
\item it has basic manipulation tools and allows to easily add new functions;
\item annotations are fast to load and modify and convenient to store;
\item annotations are stored in a format that can be manually inspected and edited;
\item it is agnostic to the format of how images are stored on disk;
\item it supports image classification, object detection, semantic segmentation, and object matching tasks in computer vision.
\end{itemize}

The toolbox, the installation instructions, the manual, and use cases are available at \texttt{https://github.com/kukuruza/shuffler}.

\section{Related work}

\subsection{Relational Model format for annotations}
Questions of data management have been studied in great detail since the early days of automated information processing. It proved convenient to store data using the Relational Model, where information is stored in a set of interconnected tables. Shortly after the Relational Model was proposed in 1970~\cite{1970_RelationalModel}, research was started on developing an appropriate language to handle it~\cite{1974_SEQUEL}, until finally the International Organization for Standardization (ISO) accepted the Structured Query Language (SQL) as a standard in 1987. Since then the Relational Model has remained dominant in industry.

Typically, a computer vision dataset can be well described by the Relational Model since a machine learning algorithm is trained on samples of training data with the same structure. In particular, in the computer vision field, an image classification algorithm can be trained on pairs $\{\text{image}, \text{label}\}$. Pairs $\{\text{image}, \text{list of bounding box}$ $\text{ coordinates}\}$ are, for example, used to train in an object detection task. Semantic pixel-wise image segmentation is trained on pairs $\{\text{image}, \text{label}$ $ \text{ map}\}$, where label map is of the same size as the image.

Storing annotations in a relational database has advantages over formats such as \text{xml} or \texttt{json}. First, the relational database is human-friendly: any SQL editor can be used to browse through the contents and to query for specific entries. Second, such database can be loaded in milliseconds, while it takes seconds to load and parse annotations in their original formats (Table~\ref{table:dataset_formats}).
Importantly, as we show in the next chapters, the Relational Model proves to be a convenient format for the purpose of building a toolbox around it because of its inherent structure and the powerful SQL language.

So far, the Relational Model has not been not popular in the modern computer vision field for several reasons. First, the Relational Model does not fit well in the academic workflow. For distributing a dataset, researchers choose universally known, human readable formats (Table~\ref{table:dataset_formats}). For training, the dataset can be serialized as files in Google \texttt{protobuf} format or similar in order to minimize the time to read the data from disk. In this work, we argue that a Relational Model is needed, not for dataset distribution or for data serialization, but rather for the intermediate step of modifying and filtering annotations, extracting a subset of the dataset, visualizing information, and other similar related tasks.

Second, different datasets define different information that needs to be stored. For example, angles are important for objects in images from traffic surveillance cameras but are irrelevant for generic images such as in the Pascal VOC dataset~\cite{PASCAL_retrospective}. That makes any particular schema hard to generalize across datasets. Instead of providing a universal schema, we focus on setting up a base that can be customized for particular needs of a particular company, group, or user. At the same time, our design of the schema is generic enough to fit the tasks of image-level classification, object detection, semantic segmentation, and object matching, which cover a large part of the computer vision landscape.



\vspace{12pt}

\subsection{Dataset management in Computer Vision}

As computer vision becomes more useful for practical applications, a number of solutions facilitating the life-cycle of a project has emerged. These solutions address different challenges of the ML pipeline.

First, some systems are designed specifically for annotating data for computer vision applications. Examples include publicly available LabelMe~\cite{2008_LabelMe}, VGG Image Annotator~\cite{2016_VGGImageAnnotator}, and CVAT\footnote{\url{https://github.com/opencv/cvat}}, as well as commercial Supervisely\footnote{\url{https://supervise.ly}}, Playmate\footnote{\url{https://playment.io}}, and Labelbox\footnote{\url{https://labelbox.com}}. These systems offer sophisticated tools for human annotators to label images in order to prepare training data for different types of machine learning tasks. Though our proposed toolbox, Shuffler, offers basic functionality for image labelling, its primary focus is processing the output of such image annotation systems.

Second, an important part of a machine learning pipeline is loading and augmenting image data. Numerous libraries, such as DALI\footnote{\url{https://github.com/NVIDIA/DALI}} released by NVIDIA, have been proposed for this task. In Figure~\ref{fig:data-preparation-pipeline}, we refer to this part of the pipeline as step $3$. In turn, Shuffler is employed on step $2$ to prepare a dataset of training data that can be further loaded and augmented during training.

Next, end-to-end product life-cycle management systems have been proposed, such as ModelHub~\cite{2017_ModelHub}, or commercial Allegro\footnote{\url{https://allegro.ai}}. These systems focus on managing experiments and trained models, while the goal of Shuffler is to provide instruments to manage training data.

As far as we know, there is no flexible, publicly available instrument to facilitate manipulating image data and their annotations in order to prepare data for a ML algorithm. The goal of Shuffler is to close this gap.

The rest of the paper is organized as follows. Section~\ref{sec:shuffler_schema} describes the schema of the SQL database, the covered use cases, and the limitations of the schema. Sections~\ref{sec:shuffler_toolbox} and~\ref{sec:shuffler_subcommands} present the Shuffler toolbox and the operations it supports. This is further explored in Section~\ref{sec:shuffler_chaining} that illustrates an important feature -- chaining multiple operations in a single command. Section~\ref{sec:shuffler_data_generators} introduces the interface to Keras and PyTorch machine learning frameworks. Section~\ref{sec:shuffler_implementation} explains adding new functionality to the toolbox. We conclude the paper in Section~\ref{sec:shuffler_conclusion}.

\section{Database schema}
\label{sec:shuffler_schema}

\begin{figure*}
  \centering
  \includegraphics[width=0.6\textwidth]{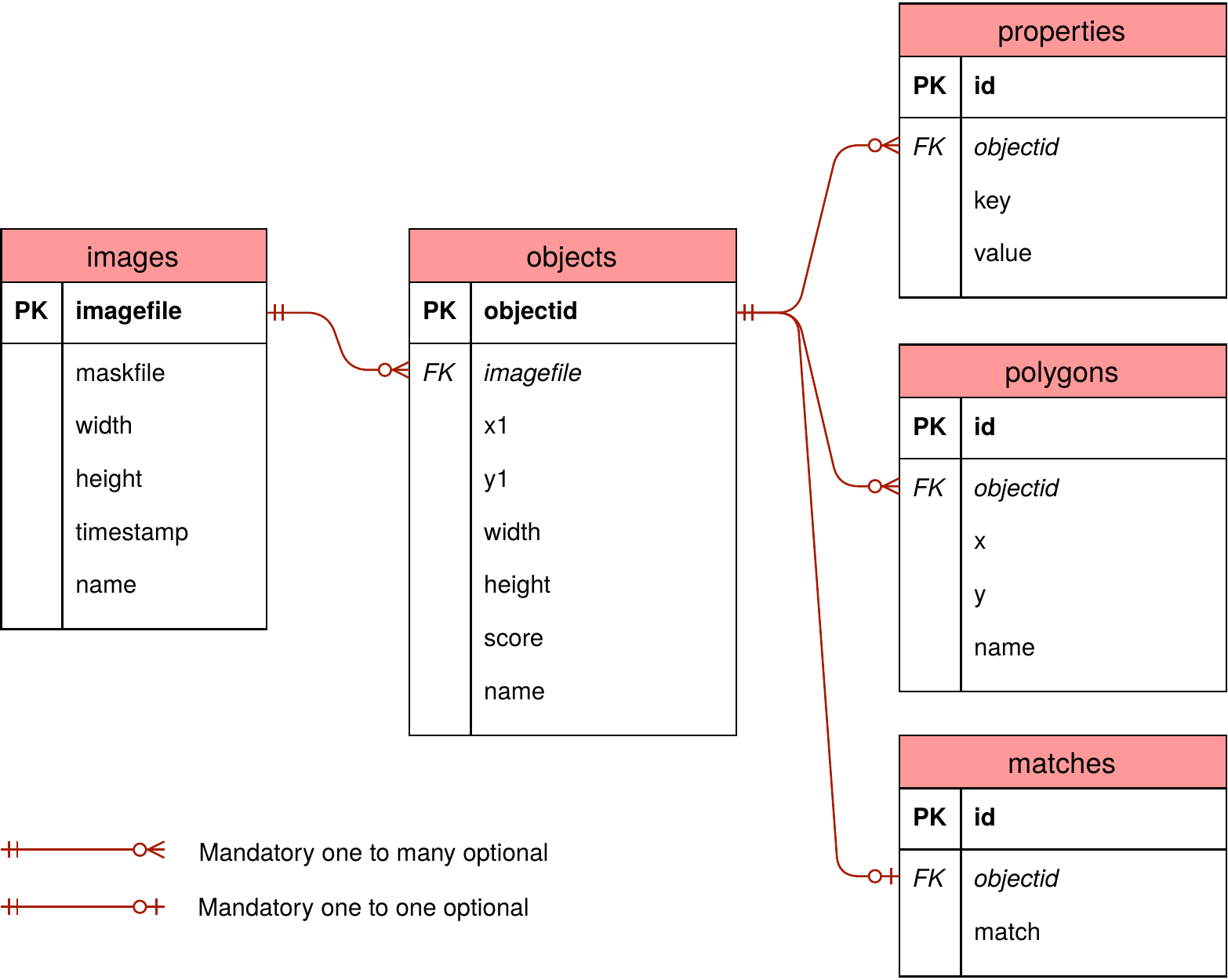}
  \caption{Database schema}
  \label{fig:shuffler_schema2D}
\end{figure*}

The core of this work is the proposed Relational Model for common datasets that were collected to train for image classification~\cite{2009_imagenet,2014_COCO}, object detection~\cite{PASCAL_retrospective, 2014_COCO, 2012_KITTI, 2008_LabelMe}, semantic segmentation~\cite{2014_COCO, 2016_bdd, 2017_Mapillary, 2018_apolloscape}, and object matching tasks~\cite{Matching_surveillance}. The proposed schema is presented in Figure~\ref{fig:shuffler_schema2D} in the form of an Entity Relation (ER) diagram. The diagram presents five tables. Each table consists of several fields. Two fields in any table are mandatory to be filled in: the unique Primary Key (PK in bold font) and the Foreign Key (FK in italic font.) 

\begin{figure*}
  \centering
  \includegraphics[width=0.58\textwidth]{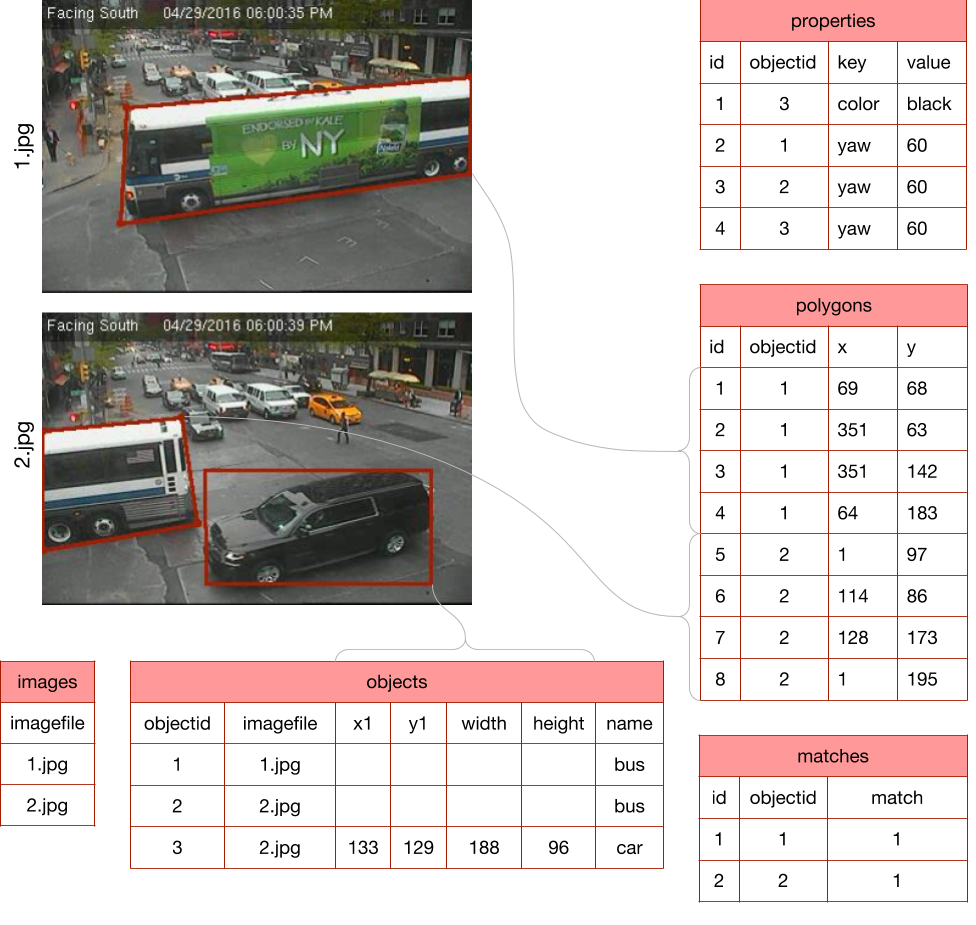}
  \caption{A database populated according to the schema in Figure~\ref{fig:shuffler_schema2D}. Only non-empty columns are shown. The same bus is recorded under \texttt{objectid=1} and \texttt{objectid=2} in the two images and is matched via \texttt{match=1}. The two bus objects are assigned a bounding polygon, while the car is assigned a bounding box.}
  \label{fig:shuffler_illustration}
\end{figure*}

In this section, we describe a typical dataset in computer vision applications and show how it corresponds to this schema. An example is provided in Figure~\ref{fig:shuffler_illustration}. It illustrates a traffic camera dataset and is focused on the tasks of object detection and matching.

Typically, a computer vision dataset consists of a number of images. Each image corresponds to an entry in \texttt{images} table and is uniquely defined by the \texttt{imagefile} field. Normally, this field contains the path to the image, but may also have other descriptors, such as the frame in a video. In the case when the dataset is focused on the image classification task, each image is associated with a label, such as ``cat'' or ``dog.'' The label is recorded in the \texttt{name} field of the \texttt{images} table. In the case when the dataset focuses on the semantic segmentation task, each image comes with a segmentation map. In this schema, the segmentation map descriptor is recorded in the \texttt{maskfile} field of the table \texttt{images}.

A dataset may contain multiple objects in each image. In this schema, each object corresponds to one entry in table \texttt{objects} and has the unique \texttt{objectid}. Each image may contain zero, one, or multiple objects, while any object must belong to some image. That is encoded as ``mandatory one to many optional'' relations between \texttt{images} and \texttt{objects} tables. In the object detection setup, each object is characterized by its bounding box, which is encoded as $\{\texttt{x}, \texttt{y}, \texttt{width}, \texttt{height}\}$ in table \texttt{objects}. Besides, each object typically belongs to a single class out of the pre-defined set of classes, for example, ``car'' and ``truck'' in Figure~\ref{fig:shuffler_illustration}. The class name, if present, is to be encoded in the \texttt{name} field of table \texttt{objects}.

Furthermore, an object may have auxiliary properties. For example, some cars in Figure~\ref{fig:shuffler_illustration} have the assigned color and the two angles, yaw and pitch, as seen by the camera. This extra information is recorded in the table \texttt{properties}, which is linked to the table \texttt{objects}. Any object may have any number of different properties, thus this schema is applicable to datasets with unstructured information on their objects.

Some datasets, such as LabelMe~\cite{2008_LabelMe}, provide fine-grained boundaries of their objects using polygons instead of rectangular bounding boxes. Our proposed schema supports this via table \texttt{polygons}. The rows in \texttt{polygons} table with the same value of \texttt{objectid} field describe the points in a closed polygon. Thus the order of their \texttt{id}'s matters: the points should be recorded either clockwise or counter-clockwise. It may happen that two polygons correspond to the same object. Then they should be differentiated with the \texttt{name} field.

Finally, the task of object matching is supported with table \texttt{matches}. The matched objects are recorded as separate entries in the \texttt{matches} table, and, as such, they have different \texttt{id} and \texttt{objectid}. However, they share the same value of the \texttt{match} field. That value uniquely identifies these particular matched objects among other matches. This idea is illustrated in Figure~\ref{fig:shuffler_illustration}.

The proposed schema comes with certain limitations. For example, it does not support the increasingly popular format of video clips~\cite{2016_bdd,2018_apolloscape}. It also does not inherently support 3D bounding boxes of objects~\cite{2012_KITTI,2018_apolloscape}, though they can be encoded with our schema via the \texttt{properties} table. In both cases the schema can be trivially extended, but such extension is beyond the scope of this work.

\section{Toolbox}
\label{sec:shuffler_toolbox}

The SQL schema alone would be useless without the tools that take advantage of it. We developed a toolbox that allows to 1) import annotations from other formats, 2) save annotations as an SQLite database, 3) modify them, and 4) export them into other formats.

A user interacts with the toolbox by executing the program \verb|shuffler.py| from the command line. In a minimal working example below, Shuffler creates a new database and prints information about it to the standard output:

\begin{lstlisting}[style=snippetstyle]
./shuffler.py printInfo

[shuffler.py:62 INFO]: will create a temporary database in memory.
=== Running printInfo ===
{'num objects': 0, 'num images': 0}
\end{lstlisting}

In this example, Shuffler calls the sub-command \texttt{printInfo}. In general, all the work with \texttt{Shuffler} is performed via sub-commands. The example below illustrates how the sub-command \texttt{importKITTI} and command-line arguments \verb|--images_dir| and \verb|--detection_dir| are used to import annotations from the KITTI dataset~\cite{2012_KITTI}. It is assumed that KITTI has been downloaded and is located in the directory \texttt{KITTI}. 

\begin{lstlisting}[style=snippetstyle]
$ ./shuffler.py importKitti
    --images_dir='KITTI/data_object_image_2/training/image_2' \
    --detection_dir='KITTI/data_object_image_2/training/label_2'
    
[shuffler.py:62 INFO]: will create a temporary database in memory.
=== Running importKitti ===
100% (7481 of 7481) |#######| Elapsed Time: 0:00:16 Time:  0:00:16
\end{lstlisting}

In this example, the database is created in-memory and is never recorded to the hard-drive. Loading and saving databases is controlled by the two command-line arguments: \texttt{-i} and \texttt{-o}. For example, KITTI annotations for the object detection task can be imported and then recorded as \texttt{kitti.db}:

\begin{lstlisting}[style=snippetstyle]
$ ./shuffler.py  -o='kitti.db'  importKitti     
    --images_dir='KITTI/data_object_image_2/training/image_2' \
    --detection_dir='KITTI/data_object_image_2/training/label_2'

[shuffler.py:36 INFO]: will create database at kitti.db
=== Running importKitti ===
100% (7481 of 7481) |#######| Elapsed Time: 0:00:17 Time:  0:00:17
[shuffler.py:122 INFO]: Committed.
\end{lstlisting}

The next example shows how to load the recorded \texttt{kitti.db} and print basic information about it:

\begin{lstlisting}[style=snippetstyle]
$ ./shuffler.py  -i='kitti.db'  printInfo

[shuffler.py:50 INFO]: will load from kitti.db, will not commit.
=== Running printInfo ===
{'image height': '4 different values',
 'image width': '4 different values',
 'matches': 0,
 'num images': 7481,
 'num masks': 0,
 'num objects': 51865,
 'properties': ['alpha', 'dim_height', 'dim_length', 'dim_width', 
 'loc_x',  'loc_y', 'loc_z', 'occluded', 'rotation_y', 'truncated']}
\end{lstlisting}

Finally, when both \texttt{-i} and \texttt{-o} are specified, a database is loaded, modified, and saved under a different name:

\begin{lstlisting}[style=snippetstyle]
$ ./shuffler.py  -i='kitti.db'  -o='clean.db'  filterObjectsAtBorder

[shuffler.py:44 INFO]: will copy database from kitti.db to clean.db.
=== Running filterObjectsAtBorder ===
100% (7481 of 7481) |#######| Elapsed Time: 0:00:03 Time:  0:00:03
[dbFilter.py:146 INFO]: Deleted 6966 out of 51865 objects.
[shuffler.py:122 INFO]: Committed.
\end{lstlisting}

The effects of all the combinations of \texttt{-i} and \texttt{-o} command-line arguments are summarized in Table~\ref{table:shuffler_rules_io}. Finally, for completeness, we present the Shuffler interface in Listing~\ref{lst:shuffler_interface}. In the next section, we focus on individual sub-commands.

\setlength{\tabcolsep}{8pt}
\begin{table*}
\caption{Different combinations of the arguments \texttt{-i} and \texttt{-o} and their meanings.}
\label{table:shuffler_rules_io}
\begin{tabular}{ccl}
\toprule
Input & Output & Description \\
\midrule
-- & -- & Create a new database in-memory. Discard it at the end. \\
\texttt{-i in.db} & -- & Open \texttt{in.db} in read-only mode. \\
-- & \texttt{-o out.db} & Create a new \texttt{out.db} and commit transactions there. \\
\texttt{-i in.db} & \texttt{-o out.db} & Open \texttt{in.db} but commit transactions to \texttt{out.db}. Backup \texttt{out.db} if it already exists. \\
\bottomrule
\end{tabular}
\end{table*}
\setlength{\tabcolsep}{1.4pt}

\begin{lstlisting}[style=snippetstyle,
  caption={Shuffler interface. Input/output is controlled by \texttt{-i/-o}, \texttt{--relpath} determines the root path that all \texttt{imagefile}-s are relative to, \texttt{logging} controls the verbosity of the output, \texttt{-h} prints out more information about arguments.},
  label={lst:shuffler_interface}
]
shuffler.py [-i IN_DB_FILE] [-o OUT_DB_FILE] 
            [--relpath RELPATH]
            [--logging {10,20,30,40}]
            [-h]
            sub-command-1 [sub-arguments-1]
            [sub-command-2 [sub-arguments-2] ...]
\end{lstlisting}

\section{Sub-commands}
\label{sec:shuffler_subcommands}

Sub-commands are the workhorse of Shuffler. Their complete list is presented in the project's official page, but can be printed out with:

\begin{lstlisting}[style=snippetstyle]
$ ./shuffler.py -h
\end{lstlisting}

Besides the global command-line arguments, each sub-command defines its own arguments. One can get help on an individual sub-command and its arguments like in the example below:

\begin{lstlisting}[style=snippetstyle]
$ ./shuffler.py printInfo -h
usage: shuffler.py printInfo [-h] [--imagedirs] [--imagerange]
Sum up and print out information in the database.
optional arguments:
  -h, --help          show this help message and exit
  --images_by_dir     print image statistics by directory
  --objects_by_image  print object statistics by directory
\end{lstlisting}

The sub-commands can be divided into several major groups:
\begin{enumerate}

\item \textbf{Import} group allows to add annotations from datasets with different formats to a new or existing database. At the moment, the functions \texttt{importPascalVoc2012}, \texttt{importKitti}, \texttt{importLabelme}, among others, are implemented.

\item \textbf{Filter} group serves to remove images or objects from the database according to some criteria. The functionality of \texttt{filterEmptyImages} and \texttt{filterObjectsAtBorder} can be inferred from their names. The function \texttt{filterObjectsSQL} is a more flexible tool that filters out images or objects based on an SQL query. For example:

\begin{lstlisting}[style=snippetstyle]
$ ./shuffler.py  -i='my.db' \
    filterObjectsSQL --where_object='width<64 AND name="car"'
\end{lstlisting}

Under the hood, this sub-command opens \texttt{my.db} and run the \texttt{DELETE} SQL query on its tables. Its simplified version for the \texttt{objects} table may look like this:

\begin{onelinerSQL}
DELETE FROM objects WHERE width<64 AND name="car"
\end{onelinerSQL}

\item \textbf{Modify} group changes entries in a database. For example, \texttt{expandBoxes} expands bounding boxes from each side, \texttt{addDatabase} merges another database with the open one, \texttt{splitDatabase} on the contrary splits the database into several parts (for example, into the train, test, and validation sets), \texttt{polygonsToBoxes} computes a bounding box for each polygon. Sub-commands in this group have unique meaning and serve various purposes. It is worth noting that all operations are performed on the database while images on disk are not modified or filtered in any way.

\item \textbf{Info} group prints aggregated information or dumps a part of the database and creates different types of plots using \texttt{matplotlib} package~\cite{Matplotlib}. For example, a histogram of angles from the ``properties'' table can be plotted as shown below:

\begin{lstlisting}[style=snippetstyle]
$ ./shuffler.py -i='my.db'  plotObjectsHistogram \
    'SELECT value FROM properties WHERE key="angle"'
\end{lstlisting}

\item \textbf{GUI} group provides a graphical interface for browsing a dataset or modifying it. A user may loop through images with bounding boxes or polygons overlaid on them using \texttt{display}, assign or change object names using \texttt{examineObjects}, and view or change matches with \texttt{examineMatches}. For example, the command below iterates over images in a random order and displays them and all the objects. We use OpenCV~\cite{opencv_library} as the backend for the graphical interface.

\begin{lstlisting}[style=snippetstyle]
$ ./shuffler.py  -i='my.db'  displayImages --shuffle --show_objects
\end{lstlisting}

\item \textbf{Evaluate} group assumes that the open database contains predictions made by a machine learning algorithm. The sub-commands evaluate these predictions with respect to another database, which contains the ground truth. Currently, evaluations of the object detection and the semantic segmentation tasks are supported. Below predictions are evaluated for the machine learning tasks.

\begin{lstlisting}[style=snippetstyle]
$ ./shuffler.py  -i='predictions.db' \
    evaluateDetection --gt_db_file='ground_truth.db'
\end{lstlisting}

\item \textbf{Export} group exports annotations to one of the supported formats as well as provides an interface to Keras~\cite{Keras} and PyTorch~\cite{PyTorch} using provided data generator classes (Section~\ref{sec:shuffler_data_generators}).

\end{enumerate}

\section{Chaining operations}
\label{sec:shuffler_chaining}

One main motivation for the toolbox design was the ability to conveniently chain operations. An example is shown in listing~\ref{lst:shuffler:chaining}. Commands are chained via the vertical bar symbol ``\texttt{|}'', that must be escaped in a Unix shell as \texttt{\symbol{92}|}, \texttt{'|'}, or \texttt{"|"}. The program exports bounding boxes of cars into a new dataset that will be further sent to LabelMe annotators. First, expand bounding boxes by 20\% from every side. Then select those cases that intersect with other cars by more than 30\%, those at the image border, those with bounding boxes of size less than 64 pixels in either dimension, and those with names other than ``van,'' ``taxi,'' or ``sedan.'' Then the bounding boxes are cropped, scaled to $64 \times 64$, and written to a new dataset -- a database with a video. Note that five calls are chained: \texttt{expandBoxes}, \texttt{filterObjectsByIntersection}, \texttt{filterObjectsAtBorder}, \texttt{filterObjectsSql}, and \texttt{cropObjects}. Without chaining, one would need to carefully store intermediate results -- one after each operation.

\begin{lstlisting}[style=snippetstyle,
  caption={Examples of chaining operations. The program crops objects into directory \texttt{crops} that will be further sent to LabelMe annotators. Note the chained calls to \texttt{expandBoxes}, \texttt{filterObjectsByIntersection}, \texttt{filterObjectsAtBorder}, \texttt{filterObjectsSql}, and \texttt{cropObjects}.},
  label={lst:shuffler:chaining}
]
./shuffler.py -i='in.db' \
  expandBoxes --expand_perc=0.2 \| \
  filterObjectsByIntersection --intersection_thresh_perc=0.3 \| \
  filterObjectsAtBorder \| \
  filterObjectsSQL --where_object='width < 64 AND name="car"' \| \
  cropObjects --edges=distort --target_width=64 --target_height=64 \
    --image_pictures_dir='crops'
\end{lstlisting}

\section{Interface to Keras and PyTorch}
\label{sec:shuffler_data_generators}

Apart from the functionality of \texttt{shuffler.py}, the toolbox also provides support for loading data in PyTorch~\cite{PyTorch} and Keras~\cite{Keras} directly from a database. Keras allows to use a custom \texttt{DataGenerator} class that loads data by batch. At the same time, a custom \texttt{ImagesDataset} class in PyTorch can be used to load individual items, which are further collected into batches by PyTorch's native \texttt{DataLoader}.

File \texttt{interface.keras.generators.py} provides custom generator classes, which can load data for the tasks of image classification, semantic segmentation, and object detection. The advantage of this class is the ability it gives a user to choose data entries from the database with \verb|where_images| and \verb|where_objects| arguments. 
Similarly, file \texttt{interface.pytorch.datasets.py} contains classes inherited from \texttt{torch.utils.data.Dataset}. Arguments \verb|where_images| and \verb|where_objects| can be used in the same way to use only a subset of the dataset. An example of using this class for semantic segmentation is shown in Listing~\ref{lst:shuffler_pytorch}.

\begin{lstlisting}[style=snippetstyle, language=Python,
  caption={We provide the class \texttt{ImageDataset} used to load data in PyTorch.},
  label={lst:shuffler_pytorch}
]
import torch.utils.data
from shuffler.interface.pytorch.datasets import ImageDataset
# Create an object of Dataset associated with 'my.db'.
dataset = ImageDataset(db_file='my.db')
# Get one item from the dataset.
image, mask = next(iter(dataset))
# The standard way to load data in PyTorch.
loader = torch.utils.data.DataLoader(dataset, batch_size=10)
\end{lstlisting}

\section{Implementation details}
\label{sec:shuffler_implementation}

The program Shuffler is implemented as a Python script, which can call one of many functions. For example, imagine a user typed:

\begin{lstlisting}[style=snippetstyle]
./shuffler.py -i my.db \ 
  filterObjectsAtBorder --border_thresh_perc 0.01
\end{lstlisting}

Shuffler will parse the first command line arguments \verb|-i my.db|, open the database \verb|my.db|, and get the ``cursor'' that allows to send queries to the database. It then will find a subparser for the function \verb|filterObjectsAtBorder|. The subparser will parse the remaining argument ``\verb|--border_thresh_perc 0.01|.'' This functionality is implemented via the \verb|argparse| package. Then Shuffler will call the function \verb|filterObjectsAtBorder| passing it the cursor and the parsed arguments.

Shuffler analyzes command line arguments sequentially and executes the sub-commands when it runs across them in the command line. That allows to chain sub-commands inside a single call to \texttt{shuffler.py}. The database is opened or created by Shuffler in the beginning according to rules~\ref{table:shuffler_rules_io}. The database cursor is passed to each function that is called by Shuffler. Therefore, each function has the possibility to perform transactions and introduce changes to the database. These changes may or may not be committed by Shuffler in the end, depending on whether the arguments \verb|-o out.db| were passed to Shuffler. 

All functions share the same interface:
\begin{verbatim}
  def myFunction(cursor, args)
\end{verbatim}
where \texttt{cursor} is an SQLite3 cursor for the open database, and \texttt{args} is the named namespace with parsed arguments for \texttt{myFunction}. That makes adding sub-commands straightforward. To add a new function and its associated sub-command, one first needs to pick a file where the function would fit the best based on its functionality. For example, all operations of filtering reside in \texttt{dbFilter.py}. Then one needs to 1) write its body, which implements the interface, 2) write the parser, and 3) register the parser in the \verb|add_parsers| function. Listing~\ref{lst:shuffler_new_function} provides the skeleton for function \texttt{myFilter}.

\begin{lstlisting}[style=snippetstyle, language=Python,
  caption={Adding a new sub-command \texttt{myFilter} to Shuffler.},
  label={lst:shuffler_new_function}
]
# Register your new parser in function add_parsers.
def add_parsers(subparsers):
  < ... >
  myFilterParser(subparsers)

# Write the parser for command-line arguments. 
def myFilterParser(subparsers):
  parser = subparsers.add_parser('myFilter',
    description='My new operation')
  parser.set_defaults(func=myFilter)
  parser.add_argument('--mandatory_arg', required=True)
  parser.add_argument('--extra_arg', type=int, default=1)
  < ... >

# Write the implementation of your function.
def myFilter (c, args):
  ''' The implementation of sub-command myFilter.
  Args:
    c:    SQLite3 cursor
    args: Command-line arguments parsed according to myFilterParser
  Returns:
    None
  '''
  < ... >
\end{lstlisting}

\section{Conclusion}
\label{sec:shuffler_conclusion}

Imagine a computer vision practitioner training a vehicle detector on the KITTI~\cite{2012_KITTI} (or another dataset) for an autonomous vehicle. As such, he/she wants to remove all labels except for ``Car,'' ``Van,'' ``Truck,'' and ``Tram,'' then to experiment if the bounding boxes around objects need to be expanded, if the objects on the image boundary should be removed, and whether it is better to also remove small objects. Furthermore, he/she would like to quickly see the distribution of objects by class and by size in the dataset. All in all, the researcher is using the workflow depicted in Figure~\ref{fig:data-preparation-pipeline}.

Annotations in the KITTI dataset come as text files, one per each image (Table~\ref{table:dataset_formats}). Making each of the dataset modifications described above requires the researcher to use KITTI's toolbox to load the data, write the custom code to filter/modify annotations, and write the annotations as the new set of text files, while somehow bookkeeping the paths to each annotation set.

The tool, Shuffler, that we developed allows to perform all these operations out of the box. Each of the original and modified annotations is stored as a SQLite database file. A researcher can directly use the provided \texttt{DataGenerator} class in Keras or \texttt{ImagesDataset} class in PyTorch during training and testing. The toolbox is made public at:
\begin{verbatim}
  https://github.com/kukuruza/shuffler
\end{verbatim}

In more general terms, we consider the workflow of a ML expert in the computer vision domain. Multiple modifications of the dataset are an important part of this workflow. No public tool or data representation is specifically designed to address this problem. In this work, we close this gap.

The design of our toolbox was motivated by the fact that data used in the computer vision field fits well the Relational Model.

The data model or the toolbox that we presented in this paper do not target the questions of the convenient distribution of the dataset for public use or the efficient data storage for fast loading by machine learning packages. Instead, we have focused data preparation, labelling, exploration, and evaluation of ML models.

%
\bibliographystyle{ACM-Reference-Format}
\bibliography{bib/related,bib/datasets,bib/technologies}

\end{document}